\documentclass[authoryear]{elsarticle}

\usepackage{amsmath,amssymb,fancyhdr}
\usepackage{accents}
\usepackage{theorem}
\usepackage{bm}
\usepackage{graphicx}
\usepackage{color,colortbl}
\usepackage[dvipsnames]{xcolor}
\usepackage{float}
\usepackage{epsfig,subfigure,pgf}
\usepackage{colortbl}
\usepackage{multirow,booktabs,array,threeparttable,multicol}
\usepackage{rotating,longtable,slashbox}
\usepackage[toc,page]{appendix}
\usepackage[authoryear]{natbib}
\usepackage{pdfpages}
\usepackage[linesnumbered]{algorithm2e}
\usepackage{pdflscape}

\usepackage{hyperref}
\usepackage{algorithmicx}

\graphicspath{{201704_results/}}

\begin{document}
\begin{frontmatter}

\title{Greedy Active Learning Algorithm for Logistic Regression Models}

\author{Hsiang-Ling Hsu}
\ead{hsuhl@nuk.edu.tw}
\author[ycc]{Yuan-Chin Ivan Chang}
\ead{ycchang@stat.sinica.edu.tw}
\author[rbc]{Ray-Bing Chen \corref{cor1}}
\ead{rbchen@mail.ncku.edu.tw} \cortext[cor1]{Corresponding author}
\address[hlh]{Institute of Statistics,
  National University of Kaohsiung}
\address[ycc]{Institute of Statistical Science,
  Academia Sinica}
\address[rbc]{Department of Statistics,
  National Cheng Kung University}

\begin{abstract}
We study a logistic model-based active learning procedure for binary
classification problems, {in which} we adopt a batch subject
selection strategy with a modified sequential experimental design
method. {Moreover, accompanying the proposed subject selection
scheme, we simultaneously conduct} a greedy variable selection
procedure {such that we can update the classification model with all
labeled training subjects.} The proposed algorithm repeatedly
performs both subject and variable selection steps until a prefixed
stopping criterion is reached. {Our numerical results show that the
proposed procedure has competitive performance, with smaller
training size and a more compact model, comparing with that of the
classifier trained with all variables and a full data set. {We also
apply the proposed procedure to a well-known wave data set
\citep{Breimanetal1984} to confirm the performance of {our}
method.}}
\end{abstract}

\begin{keyword}
  Active learning algorithm \sep $D$-efficiency criterion \sep forward selection \sep graft optimization
\end{keyword}
\end{frontmatter}

\section{Introduction}
To train a classification model, labeled data are essential when a
training/testing framework is adopted, and its classification
performance relies on both the size and quality of the training
subjects used for learning. In a Big Data scenario, we might easily
meet a huge size data set; however, the labeled information may be
limited in it, and an abundance of unlabeled subjects are available.
{To prevent money laundering, \citet{Dengetal2009} studied the
method for building a detection model using bank account data. This
is a good example of the situation because in this situation the
label of interest (money laundering account) will be limited in a
regular bank account data set.}  It would require a huge amount of
time and resources to verify whether an account is suspicious or
non-suspicious, even though the major parts of the transactions in a
bank account should be normal. Efficiently determining the potential
risks within a bank account in addition to effectively and
efficiently using the unlabeled subjects to improve the
classification rule is the key issue, and the concept of active
learning can be applied to this situation.

When we train a classifier in an active learning manner, we need to
annotate the unlabeled data and recruit them into the training set,
which can be done with the information of a model built on the
labeled data at the current stage. In the literature, it is pointed
out that people can usually learn the a satisfactory model
economically with such a procedure \citep{Cohn94a, settles11,
Settle2012}. There are many classification performance indexes, and
it is clear that this subject selection process may depend on the
targeted index \citep{settles.tr09, hsu10, settles11}. For example,
\cite{Culver06} studied active learning procedures that maximize the
area under the ROC curve (AUC), \cite{Long2010} were interested in
the ranking of the data, and \citet{Dengetal2009} used an active
learning study focusing on accuracy via experimental design.

When there are many redundant variables (predictors) in the
classification model, the model tends to over-fit the training
subjects, which also increases prediction uncertainty. Thus,
identifying a compact classification model is helpful in terms of
prediction. 
Since the question is how to efficiently select the most informative
subjects to join the training set, in this paper, we use active
learning approaches to address binary classification problems using
logistic models, and modify the subject selection (query) approach
used in \citet{Dengetal2009} to a batch sampling procedure, which
will make our procedure suitable for big data scenarios. We include
a variable selection {step} in our procedure, in addition to the
subject selection scheme, for systematically improving the
prediction ability and avoiding the over-fitting phenomena of the
final classification model.

We organize the rest of this paper as follows. Section 2 presents
the details of the subject selection and variable selection steps,
and then we propose an active learning algorithm both features.
Numerical results are given in Section 3. In addition to the
simulation studies, we apply our algorithm to a well-known wave data
set used in \cite{Breimanetal1984}. We present a brief discussion
and conclusion in Section 4.

\section{Methodology}

We consider a pool-based active learning procedure as studied in
\cite{Lewis94} and {assume that to obtain those unlabeled data is
cheap and to query their label information is  expensive}. Hence, we
should rationally select the unlabeled subjects from this large pool
for being labeled to reduce the overall cost of model learning. We
state the general framework of the pool-based active learning
methods as Algorithm \ref{alg:Framwork}.
\begin{algorithm}[h]
\caption{A General Framework of Pool-based Active Learning
Algorithm} \label{alg:Framwork}
    {\bf Initialization:} An initial labeled training set and a pool of unlabeled data\\
\Repeat{The stopping criterion is satisfied}{
   {\bf Learn} the current model based on the current labeled training data.\\
   {\bf Select} points from unlabeled set via a query strategy
framework based on the current learned model.\\
   {\bf Query} labels for these selected points and update the
training set.}
\end{algorithm}

As mentioned before, to learn a classifier by including the entire variables because it would increase the prediction errors and more data are necessary to construct a stable classifier, and it is especially the case, when we make a sparse model assumption, a compact classification model is preferred.
Thus, besides the subject selection, but the variable selection procedure is our another goal in our learning process.


The {variable} selection frame has two common approaches: forward
selection and backward elimination. In our study, we use the forward
selection scheme that increases the size of the {variable} set by
adding a new variable to the current model at a time, and a greedy
selection approach is adopted. We employ a {variable} selection
framework to effectively reduce the number of {variable} with a
heuristic procedure {\citep{whiteny1971}}. Based on the characters
discussed above, we propose an active learning algorithm, which
integrates both batch-subject selection and greedy variable
selection feature together, and we will refer to this greedy active
learning algorithm as {\it GATE} throughout the rest of this paper.
Basically, we add a variable selection step once we have an updated
training set, and in each iteration of {\it GATE}, we will add more
labeled samples, and re-justify our classification model.

\subsection{Logistic Model for Binary Classification}

Let $\Xi_S$ denote the index set of the whole sample points and $\Xi_s$ be the current training index set with labeled data. Thus, ${\Xi^c_s=}\Xi_S \setminus \Xi_s$ is the pool of the unlabeled data.
In this section, we focus on how to identify the batch unlabeled subjects from $\Xi^c_s$ for binary classification based on a logistic model, and meanwhile,  propose a two-stage query procedure by putting the uncertainty sampling with the optimal design criterion together. Afterward, we introduce a greedy forward selection to update the current model by selecting a candidate variable from ${\Xi^c_v=}\Xi_V \setminus \Xi_v$,
where $\Xi_V$ is the index set of the whole variables,
and $\Xi_v$ denotes the index set of the current active variables in the logistic model.

Assume that the $i$th individual variate $Y_i \in \{0, 1\}$ is
binary variable with
$$P(Y_i=1)=p_i=E(Y_i)\mbox{ and } P(Y_i=0)=1-p_i,$$
and let the feature values of the $i$th subject, ${\bf
x}_{i,p}=({x_{i,j}})^{\top},\, i\in \Xi_{s},\, j\in \Xi_{v}$.
Suppose that the number of the dimension of ${\bf x}_{i,p}$ is equal
to $p$. Then we can fit this data set with a logistic regression
model below :
$$p_i=F({{\bf x}_{i,p}}|\bm{\beta}_p)=\frac{\exp\{\bm{\beta}^{\top}_{p}{\bf x}_{i,p}\}}{1+\exp\{\bm{\beta}^{\top}_p {\bf x}_{i,p}\}},$$
where $\bm{\beta}_p$ is a $p \times 1$ unknown parameter vector.
The log-likelihood function of the logistic model using the {labeled} (training) set, $\{(Y_i, \mathbf{x}_{i,p}), i \in \Xi_s\}$, is
$$L= \sum_{i \in \Xi_s}\left\{Y_i\ln F({\bf x}_{i,p}|\bm{\beta}_{p})+(1-Y_i)\ln (1-F({\bf x}_{i,p}|\bm{\beta}_{p}))\right\}.$$
Thus, we can use the {maximum} likelihood estimation (MLE) to estimate $\bm{\beta}_p$.
It is known that there is no close-form expression in the MLE approach,  and the numerical optimization approach, like Newton's method and iteratively re-weighted least squares (IRLS), are commonly used in this case.
Once we obtain the estimate of $\bm{\beta}_p$,  $\hat{\bm{\beta}}_p$,
we can predict the label of the $i$th observation by
$$\hat{Y}_i=1 \mbox{ if }\hat{p}_i=\frac{\exp\{\hat{\bm{\beta}}_{p}^{\top}{\bf x}_{i,p}\}}{1+\exp\{\hat{\bm{\beta}}_{p}^{\top}{\bf x}_{i,p}\}} {>\alpha},$$ with the pre-specified value $\alpha$, say, for example, $\alpha = 0.5$.

\subsection{Evaluation of subjects via an optimal design criterion}

If we use the experimental design methodologies properly to query the next points, the information concealed in a large data set will be extracted quickly.
Using some well-developed techniques in design theories \citep[see][]{Atkinson96, Fedorov1972}, we can effectively select samples, and some analytic results of optimal designs for parameter estimation in logistic models have been derived.
In the active learning literature, there are already many optimal design-based active learning approaches for recruiting new samples into training sets \citep[for example, see][]{Cohn94, Cohn96, Dengetal2009}.
\citep[For general information about the optimal design theory, please refer to][]{Silvey80}.
Because, in our current problem, we only have unlabeled samples instead of a compact design space as in the conventional design problems, it is hard to apply these analytic designs to our problem; especially, when we consider the variable selection as a part in an active learning process. How to quickly locate data points that are close to the analytic ones among a huge data set
will be the main issue when applying the design criteria to active learning processes.
Thus, it becomes a computational problem, instead of construction problem, in this area.

Suppose $\{\mathbf{x}_{i, p} = ({x}_{i,j})^{\top}, i \in \Xi_{s},\, j\in \Xi_{v}\}$ denotes the current labeled point set of size $n$ with a variable length equal to $p$. Following the definition in the optimal design \citep{Silvey80}, we set a
design $\xi_{n,p}$ at points ${\bf x}_{i,p},\,i\in \Xi_{s}$ with equal weights $1/n$.
Given the parameter estimate, $\hat{\bm{\beta}}_p$, the information matrix of the logistic model is
\begin{align}
M(\xi_{n,p},
\hat{\bm{\beta}}_p)=\bm{X}^{\top}_{n,p}\bm{W}_{\hat{F},p}\bm{X}_{n,p}/n,
\label{eq:infM}
\end{align}
where $\bm{X}_{n,p}$ is the $n \times p$ design matrix with ${\bf x}_{i,p}$ as its $i$th row, and $\bm{W}_{\hat{F},p}$ is {an} $n \times n$ diagonal matrix with the $i$th diagonal element $w_{ii}$ equal to $$w_{ii}=\hat{F}({\bf x}_{i,p}|\hat{\bm{\beta}}_p)[1-\hat{F}({\bf x}_{i,p}|\hat{\bm{\beta}}_p)],\, i \in \Xi_s.$$
It is clear that the information matrix in (\ref{eq:infM}) depends on the current parameter estimate, and  therefore, a ``locally optimal'' criterion will be used for subject selection consideration \citep{Silvey80}.
Suppose ${\bf x}_{t,p}=(x_{t,j},\,j\in \Xi_v)^{\top},\, t\in \Xi_{s}^c$ is an unlabeled subject to be added to the design $\xi_{n,p}$, then following \citet{Fedorov1972}, the $(n+1)$-points design including ${\bf x}_{t,p}$ is $\xi_{n+1,p}=\frac{1}{n+1}\bar{\xi}_{t,p}+\frac{n}{n+1}\xi_{n,p}$, where $\bar{\xi}_{t,p}$ is a design that puts all of the mass at the point ${\bf x}_{t,p}$.
Therefore, $\xi_{n+1,p}$ is equally supported on the points $\{{\bf x}_{i,p}, i\in \Xi_{s}\} \cup \{{\bf x}_{t,p}\}$.
We then use the efficiency of $\xi_{n+1,p}$ based on $\xi_{n,p}$ via the relative $D$-efficiency among the corresponding information matrices,
       \begin{align}
         \mbox{reDeff}({\bf x}_{t,p})=\frac{|M(\xi_{n+1,p},\hat{\bm{\beta}}_p)|^{1/p}-|M(\xi_{n,p},\hat{\bm{\beta}}_p)|^{1/p}}{|M(\xi_{n,p}, \hat{\bm{\beta}}_p)|^{1/p}},\label{eff}
       \end{align}
to measure the effectiveness of the new subject,
%
%
and  we want to select the next point, $\mathbf{x}^*$,
\begin{align}
     {\bf x}^*=\underset{\{{\bf x}_{t,p}, t\in \Xi_s^c\}}{\mbox{arg}\max}\, \mbox{reDeff}({\bf x}_{t,p}), \label{query_eq}
\end{align}
which maximizes (\ref{eff}) among all points $\mathbf{x}_{t,p},\,
t\in \Xi_{s}^c$ based on the  labeled training set and current
logistic model. {Because
 $${\bf x}^*=\underset{\{{\bf x}_{t,p}, t\in \Xi_s^c\}}{\mbox{arg}\max}\, \mbox{reDeff}({\bf x}_{t,p}) = \underset{\{{\bf x}_{t,p}, t\in \Xi_s^c\}}{\mbox{arg}\max}\,
 |M(\xi_{n+1,p}, \hat{\bm{\beta}}_p)|,$$
to select a point satisfying \eqref{query_eq} is equivalent to that
in the locally $D$-optimal criterion.}

\subsection{{Two-stage} query procedure}
Because a complete search is exhausted and computationally inefficient, when the size of the unlabeled data is huge and the uncertainty sampling strategy can reduce the searching time, the idea of uncertainty sampling is popularly used in many active learning processes in the literature.
Here, we also adopt this strategy, and proposed a two-stage procedure.
Before applying the methods of experimental design, we will first identify a candidate set based on the current logistic model with a pre-specified threshold value $\alpha$ as follows.
For each unlabeled point $\bf{x}$, we define $d({\bf x}|\hat{\bm{\beta}})=|\hat{F}({\bf x}|\hat{\bm{\beta}})-\alpha|$ to measure uncertainty with respect to the current model, $\hat{F}(\cdot|\hat{\bm{\beta}})$.
We could encompass the uncertainty candidate subjects from the unlabeled data set $\{{\bf x}_{t,p}, t \in \Xi_{s}^c\}$, i.e.,
  \begin{align}
  \{{\bf \tilde{x}}\}=
  \{{\bf x}_{t,p}: d({\bf x}_{t,p}|\hat{\bm{\beta}})\leq d_0, t\in \Xi_{s}^c\}, \label{eq:CTH}
  \end{align}
where $d_0$ is a pre-specified constant to determine the scope of
the pool of the candidate subjects. In this paper, we suggest
setting $d_0=d_{(h)}$, where $h$ is a given {integer} and $d_{(j)}$,
$j=1,\ldots,H$, are the distinct order statistic values of $\{d({\bf
x}_{t,p}|\hat{\bm{\beta}}),\,t\in\Xi^c_s\}$.

If the threshold hyper-plane can be estimated accurately, then we
can locate the targeted subjects efficiently. To extract the
concealed information in these subjects, $\{\bf \tilde{x}\}$, we
select the next labeled point $\mathbf{x}^*$ as the one maximizing
the the relative $D$-efficiency {in Eq. (\ref{query_eq}).} Moreover,
if we choose an $h$ equal to the largest integer $H$, then $\{
\mathbf{\tilde{x}}\} = \{\mathbf{x}_{t,p}, t \in \Xi_{s}^{c}\}$,
then in this case, the {two-stage} procedure is the same as the
locally $D$-optimal approach. When there is only one element in
$\{\bf \tilde{x}\}$, the proposed query approach is equivalent to
the uncertainty sampling.

\noindent{Remark:}
For given $\alpha \in (0, 1)$,  the decision boundary of a logistic model can be defined as
$ l_{\alpha}({\bf x})=\{{\bf x}: F({\bf x}|\bm{\beta})=\alpha\}. $
\citet{Dengetal2009} treated a binary classification as to obtain separation boundary estimation problem, thus they chose a few candidates close to the estimated decision boundary based on the current learning model, and then selecting their next sample using a locally $D$-optimal criterion.  Their two-stage procedure integrates the concept of the uncertainty sampling \citep{Lewis94} and an optimal design method.

\subsection{Grafting technique for greedy selection procedure}

To fit a logistic regression model with a large number of variables and too many redundant variables would cause some computational difficulties in parameter estimation, and enlarge the prediction variation.
Since to have a large number of variables, $P$, is common in this big data era,  we want to identify a compact model for the binary classifier and under such a sparse model situation.
In this paper, we adopt the concept of greedy forward selection algorithm is used for variable selection from computational consideration.  In fact, for reducing the computational cost, \citet{Efron2016} also suggests using a forward selection approach to identify a proper classification model.

\paragraph{Single Feature Optimization (SFO) procedure:}
\citet{singhetal2009} introduced a greedy-type feature selection procedure for logistic regression models.
Suppose $x_p$ is a candidate variable and in the current logistic model, we do have $x_1, \dots, x_{p-1}$. Then, in SFO, instead of re-estimating the coefficient vector $\bm{\beta}_p =(\beta_1, \dots, \beta_{p})$, it learns an approximate model by fixing the original parameters for $x_1, \dots, x_{p-1}$ and optimizing the parameter of the new variable, $\beta_p$ via the log-likelihood function $L$ with respect to $x_p$, i.e.,
      $$\hat{\beta}_p = \underset{\beta_p}{\mbox{arg}\max}~ L.$$
By this method, merely $P -(p-1)$ approximate models need to be created at each iteration of forward selection. The estimation value of $\beta_p$ is computed according to Newton's method.
To identify the next added variable,
\citet{singhetal2009} proposed to score the new feature variable $x_{p}$ by evaluating the approximate model with a proper evaluating index, like AIC or prediction error.

Instead of evaluating the variable effect based on the approximation of parameter estimation, \citet{Perkinsetal2003} proposed another greedy forward selection approach, called the grafting technique, based on the gradient of the log-likelihood function for the newly added variable. With fixed parameters $\beta_1, \dots, \beta_{p-1}$, the variable with the largest magnitude of gradient is added to the model, i.e.,
        \begin{align}
          \underset{p \in \Xi_v^c}{\arg\max}\left\vert\frac{\partial L}{\partial \beta_p}\right\vert=\underset{p \in \Xi_v^c}{\arg\max}\left\vert\sum_{i\in \Xi_s} x_{i,p}(Y_i-p_i)\right\vert.\label{gradient}
        \end{align}
Note that based on (\ref{gradient}), the grafting technique is similar to the matching pursuit \citep{Mallat_93} (or weak greedy algorithm) used in the variable selection for the regression model, and in this greedy forward selection algorithm, we only need to compute the inner product operator of $\mathbf{x}_p$ and the response vector.
\paragraph{GLMNET:}
L1 regularized logistic regression is a lasso-type method and can also be used for variable selection.
GLMNET \citep{Friedman10} is a Newton-type algorithm to identify the corresponding L1 regularized parameter estimates for logistic models.

\noindent{Remarks on the differences between the grafting technique and lasso approach}:
The following remarks state some differences between the grafting technique and lasso approach, which are also commonly discussed in the literature.
\begin{enumerate}
    \item[1.]  The grafting technique is a sequential approach that adds one variable each time. Thus, when we start from a null model, we only have a small model in our active algorithm.
    \item[2.]  Consider the computational complexity of the selection approach. To add one variable via the grafting technique, it would take $O(nP)$, and it is $M \times O(nP)$ if we add $M$ variables to the model. According to Table 2 in \citet{Yuan:2012}, GLMNET would take $O(nP)$ for one sweep of $P$ variables, and overall complexity would be $M_o \times (O(nP) + M_i \times O(nP)),$ where $M_o$ and $M_i$ are the number of iterations for the outer and inner loops, respectively. Thus, the grafting technique should have an advantage in computational cost if we add few variables, i.e., $M < M_o$.
    \item[3.] According to \citet{kubicaaetal2011}, the grafting technique is easily distributed to take advantage of parallel computing techniques. We can divide variables into several disjoint variable sets and then compute inner product values in (\ref{gradient}) individually. Finally, return these values to the master to identify the next added variable with the largest magnitude value.
\end{enumerate}

\subsection{Main Algorithm}

The proposed {\it GATE} algorithm has two parts: (1) identify the proper subjects for labeling, and (2) find a compact classification model.
Because to identify the proper variable to be added at each iteration, additional information will be required, and to rely it on just only one newly labeled point should not be enough.
Thus, we will include a batch of size $n_q$ instead, and we believe that this is appropriate for the most big data applications.
Of course, the batch size $n_q$ can be a tuning parameter in {\it GATE} algorithm, and may vary according to applications.

Our query procedure relies on the $D$-optimal criterion, hence we will stop our learning algorithm when the difference between the relative $D$-deficiencies is small enough, which is stated as follows:
\begin{eqnarray*}
\frac{\left| |M(\xi_0,\hat{\bm{\beta}}_{k})|^{1/k} - |M(\xi_1,\hat{\bm{\beta}}_{k+1})|^{1/(k+1)}\right|}{|M(\xi_0,\hat{\bm{\beta}}_{k})|^{1/k}} < \varepsilon,
\end{eqnarray*}
The constant $\varepsilon$ is a pre-specified threshold value and
$\xi_0, \xi_1$ {represent} the designs for the subjects with respect
to the $k$ and $k+1$ selected variables respectively. Hence, we stop
the {\it GATE} algorithm when the design with the additional
subjects cannot significantly increase the information. The detailed
of each step of our {\it GATE} method is now stated below as
Algorithm~\ref{alg:gate}.
\begin{algorithm}[h]
{\small
\caption{Active Learning with Batch Sampling and Forward Variable Selection Method} \label{alg:gate}

{\bf Initialization}: Learn the current logistic classifier model by estimating $\bm{\beta}_0$ based on the initial labeled set with $n_0$ points\;
\emph{Let $k = \# \Xi_v$ and set Crit = 1}\;
\While{ Crit $\geq \varepsilon\, \&\, k\leq P$ }{
  Estimating $\hat{\bm{\beta}}_k^{(0)} =(\hat{\beta}_v,\,v\in\Xi_v)^{\top}$ based $\{{\bf x}_{i,v}, \, i\in \Xi_{s},\, v\in \Xi_{v}$\}\;
  (\emph{batch active subject learning})\;
  \For{$1\leq t\leq n_q$}{
    \For{$j\in {\Xi_{s}^c}$}{
      Calculate $d_j=|\hat{F}({\bf x}_{j,k}|\hat{\bm{\beta}}_k^{(t-1)})-\alpha|$\;
     }
     Set $d_0=d_{(h)}$ as the $h$th order statistic of $d_j$ and $\Xi_I = \{i| i \in {\Xi_{s}^c}, d_i \leq d_{0}\}$\;
     Identify the point ${\bf x}_{i_t,v}=\underset{\{{\bf x}_{t,v}, t\in \Xi_I\}}{\mbox{arg}\max}\, \mbox{reDeff}({\bf
     x}_{t,v})$\;
     Query ${\bf x}_{i_t,v}$ and denoting $Y_{i_t}$ as its label\;
     Update $\Xi_{s}=\Xi_{s} \cup \{i_t\}$. Re-estimating $\hat{\bm{\beta}}_k^{(t)}$ based $\{{\bf x}_{i,v}, \, i\in \Xi_{s},\, v\in \Xi_{v}$\}\;
    }
    Define $\hat{\bm{\beta}}_k^* = \hat{\bm{\beta}}_k^{({n_q})}$\;
    Compute $M_0=|M(\xi_0,\hat{\bm{\beta}}_k^*)|^{1/k}$, where $\xi_0$ equally supports on $\{{\bf x}_{i,v},\, i\in \Xi_{s},\, v\in \Xi_{v}\}$
    and $\hat{p}_i=\hat{F}({\bf x}_{i,v}|\hat{\bm{\beta}}_k^*)$\;
 (\emph{variable selecting})\;
 \For{$u \in \Xi_{v}^c$}{
  Compute $g_u= \Big|\sum_{i\in\Xi_{s}} x_{i,u}(Y_i-\hat{p}_i)\Big|$\;
 }
  Select $u^*=\max_{u\in\Xi_{v}^c}g_u$ and update $\Xi_{v'}=\Xi_{v} \cup \{u^*\}$\;
  Re-estimate $\hat{\bm{\beta}}_{k+1}^*$\;
  Obtain $M_1=|M(\xi_1,\hat{\bm{\beta}}_{k+1}^*)|^{1/(k+1)}$ where $\xi_1$ is equally supports on $\{{\bf x}_{i,v'},\, i\in \Xi_{s},\, v'\in \Xi_{v'}\}$\;
  Compute {\it Crit}= $\left|{M_1-M_0}\right|/M_0$\;
  \eIf{Crit $< \varepsilon$}{
   $\Xi_{v}=\Xi_{v'}\setminus\{u^*\}$\;
   }{
   Updating $\Xi_{v}=\Xi_{v'}$ and $k = k + 1$\;
  }
  }
 Final re-estimate $\hat{\bm{\beta}}^*$ based on $\{{\bf x}_{i,v},\, i\in \Xi_{s},\, v\in \Xi_{v}\}$ with the selected training data set\;
 Estimate $\hat{F}({\bf x}_{j,v}|\hat{\bm{\beta}}^*)$ with ${\bf x}_{j,v}=(x_{j,v},\,v\in\Xi_v)^{\top}$ for all $j$ in the testing data set\;
 Obtaining the estimated labels for the testing data set, that is, $\hat{Y}_j^t=1$ when $\hat{F}({\bf x}_{j,v}|\hat{\bm{\beta}}^*)>\alpha$ \;

}
\end{algorithm}

\section{Simulation Studies and an Artificial Example}

We conduct several simulation studies to illustrate the performance of the proposed algorithm {\it GATE} in terms of the classification rate based on the training sample size used, and the variable selection status as well using both the synthesized data and a well-known wave data set from \citep{Breimanetal1984}.

\subsection{Simulations}

For the synthesized data, we generate the response, $Y_i$'s, from Bernoulli distribution with the probability
$$p_i=\frac{\exp\{\bm{\beta}^{\top}{\bf x}_i\}}{1+\exp\{\bm{\beta}^{\top}{\bf x}_i\}},$$
where $\mathbf{x}_i$ is a $P \times 1$ predictor vector and $\bm{\beta}$ is the corresponding parameter vector. For the $i$th predictor vector $\mathbf{x}_i = (x_{i,1}, \dots, x_{i, P})^{\top}$, we fix $x_{i,1}$ as $1$ and $x_{i,j}, j > 1$ are independently generated from a normal distribution with mean $\mu_j$ and unit standard deviation, where $\mu_j$ is assumed from the uniform distribution within $(-1,1)$. Here, $P$ is set as 100.
Moreover, due to the sparse model assumption, we do have several different scenarios for the parameter vectors.
\begin{enumerate}
     \item[] Case 1: The first five parameters are chosen as follows,
        $(\beta_1,\beta_2,\ldots,\beta_{5})^{\top}=( 0.5,-2.0, -0.6, 0.5,  1.2),$
        and the other $\beta_i, i > 5,$ are set as zeros.
     \item[] Case 2:
     The parameter vector is
        $(\beta_1,\beta_2,\ldots,\beta_{5})^{\top}$ $=(5, -20,  -6,   5,  12)$, which is equal to 10 times of the vector in Case 1.
     \item[] Case 3:
        The number of the nonzero parameters are 6 and these 6 parameters are set as
        $(\beta_1,\beta_2,\ldots,\beta_{6})^{\top}=( 1,-4,-2,2,3,7).$
        The other parameters are fixed as zeros.
\end{enumerate}
Note that 20,000 independent samples are generated for each logistic regression model, and the sample size of the testing data set is $N_v= 5,000$.

We repeat the simulation 1,000 times for each case by independently regenerating the 20,000 samples.
To implement the proposed active learning approach, we randomly select $n_0 = 100$ points collected as the initial subject set from the training set, and assume the other training points are unlabeled.
At each iteration, we set $h=200$ for the scope for the pool of the candidate subjects and the batch size $n_q = 30$.
Thus, we sequentially select 30 points based on the two-stage query procedure.
Since we assume no prior knowledge for the samples, we set the threshold value, $\alpha$, in the logistic classification equal to 0.5; that is,
$$\hat{Y}_i = 1 \mbox{ if } \hat{p}_i = \frac{\exp\{\hat{\bm{\beta}}^{\top} \mathbf{x}_i\}}{1 +\exp\{\hat{\bm{\beta}}^{\top} \mathbf{x}_i\}} > \alpha = 0.5.$$
We then set $\varepsilon=10^{-2}$ for our stopping criterion.

We use both Accuracy (ACC) and Area Under the Curve (AUC) of the receiver operating characteristic (ROC) to evaluate the performance in all cases.
Both measurements are targeted on the fraction of all instances that are correctly categorized, and thus they are larger-the-better characteristics.
To illustrate the selection performance, we consider the True Positive Rate (TPR), the False Positive Rate (FPR), and both are defined as follows.
Let $\mathcal{J}$ be the set of the true active variables
and $J$ be the set of the variables identified via the considered learning procedure, i.e.,
\begin{eqnarray*}
 \mathcal{J}=\{j: \beta_j\neq 0\}, \mbox{and }  J=\{k: \hat{\beta}_k\neq 0\}.
\end{eqnarray*}
Then,
\begin{align*}
\mbox{TPR}&=\frac{\mbox{number of } \{J\cap \mathcal{J}\}}{\mbox{number of } \{\mathcal{J}\}};\\
\mbox{FPR}&=\frac{\mbox{number of } \{J\cap \mathcal{J}^c\}}{\mbox{number of } \{\mathcal{J}^c\}}.
\end{align*}
TPR is the rate of active variables identified correctly, and FPR is the rate of inactive variables that are included in the model. Therefore, a large value of TPR indicates better performance, whereas a smaller value of FPR indicates a better performance than the larger value.

Table \ref{sim_table1} summarizes the results of the 1,000 simulations with respect to three different parameter cases.
Among these three cases, the first case should be the one with the worst classification rate due to the small non-zero parameter values. As shown in Table \ref{sim_table1}, both classification measurements of Case 1, ACC and AUC are still larger than 0.82, and on average, we only take around 530 points from the whole candidate set, which contains 15,000 points. The TPR values to evaluate the variable selection performance for these three cases are higher than 0.94. This means that most of the true active variables are included in the final model; however, over-selection problems do exist due to the potential weakness of the greedy forward selection algorithm. Finally, without taking the labeling cost into account, the proposed algorithm is quite efficient because the average CPU times are less than 4.5 minutes.

\begin{figure}[t]
   \subfigure[Case 1: Variable Selection Frequencies for GATE ]{\includegraphics[width=1\textwidth]{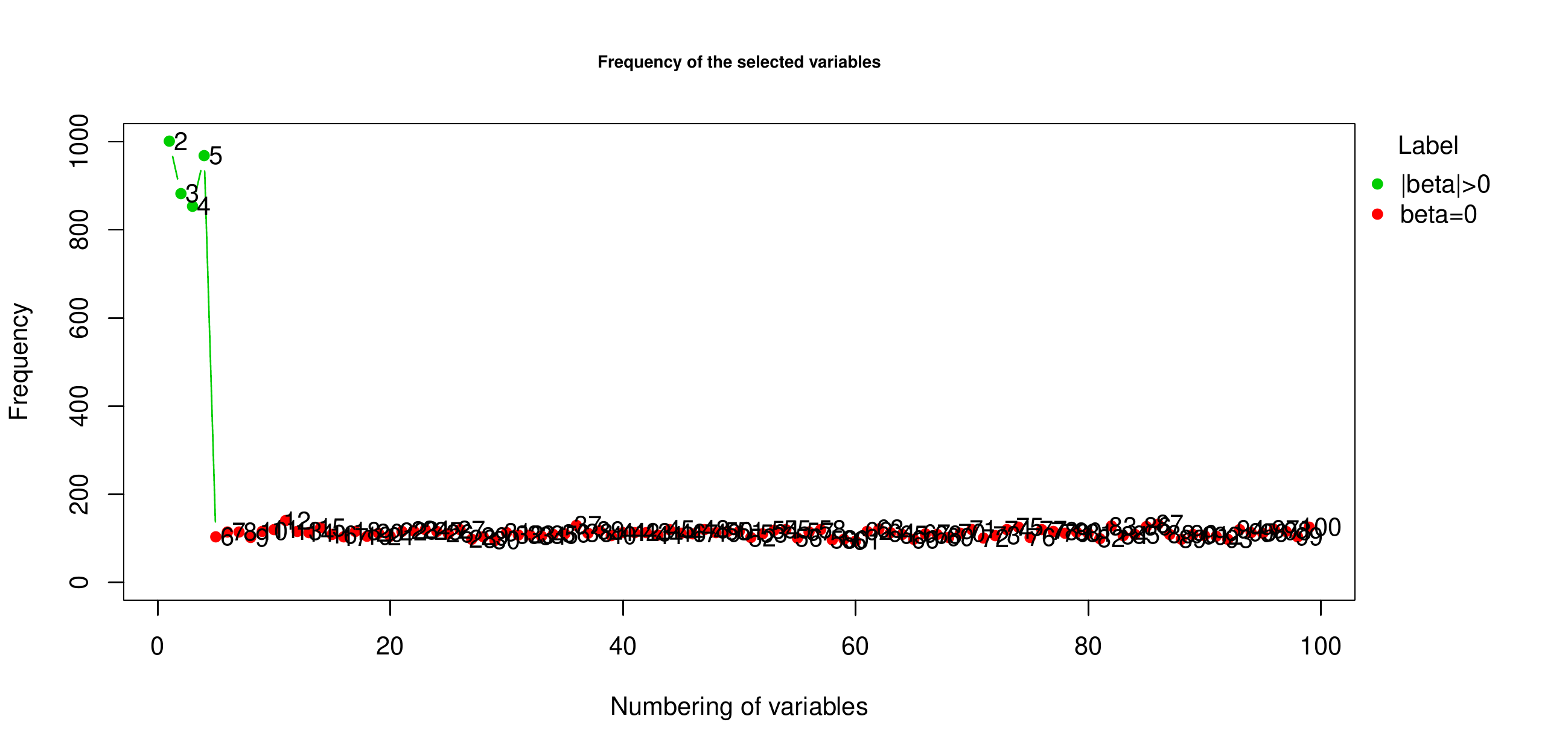}}\vspace*{-0.3cm}

   \subfigure[Case 1: ROC curve for (A)]{\includegraphics[width=0.45\textwidth,trim={0 0 0 0.5cm},clip]{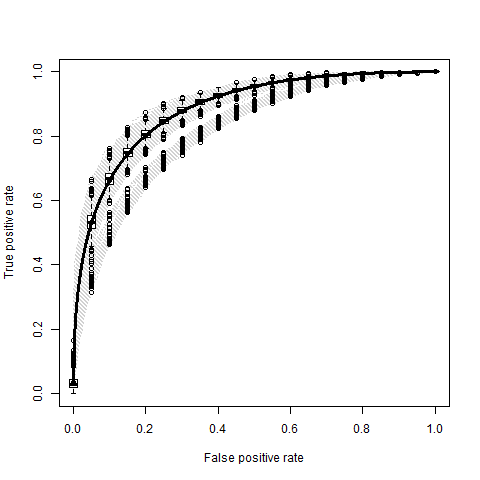}}\vspace*{-0.2cm}
   \subfigure[Case 1: ROC curve for (B)]{\includegraphics[width=0.45\textwidth,trim={0 0 0 0.5cm},clip]{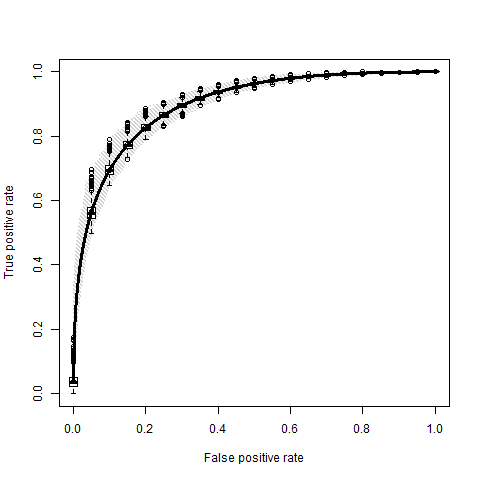}}\vspace*{-0.2cm}

   \subfigure[Case 1: ROC curve for (C)]{\includegraphics[width=0.45\textwidth,trim={0 0 0 0.5cm},clip]{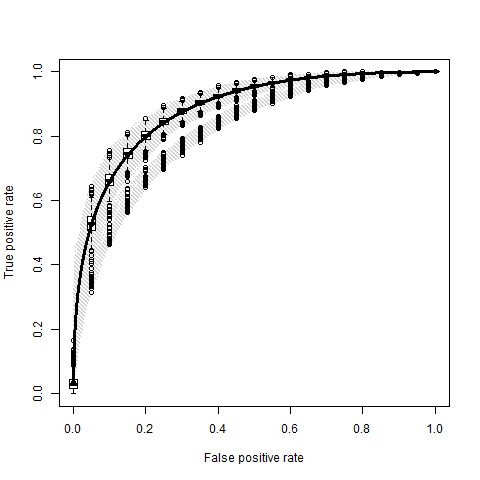}}\vspace*{-0.2cm}
   \subfigure[Case 1: ROC curve for (D)]{\includegraphics[width=0.45\textwidth,trim={0 0 0 0.5cm},clip]{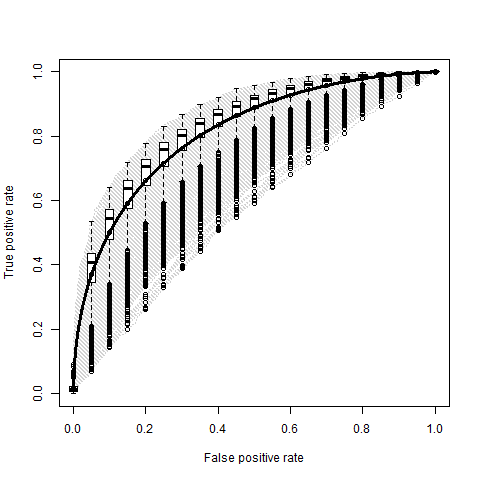}}\vspace*{-0.2cm}
   \caption{The performances of the different approaches for Case 1 with 1000 replications}\label{case1_sf}
\end{figure}

\begin{figure}[t]
   \subfigure[Case 2: Variable Selection Frequencies for GATE]{\includegraphics[width=1\textwidth]{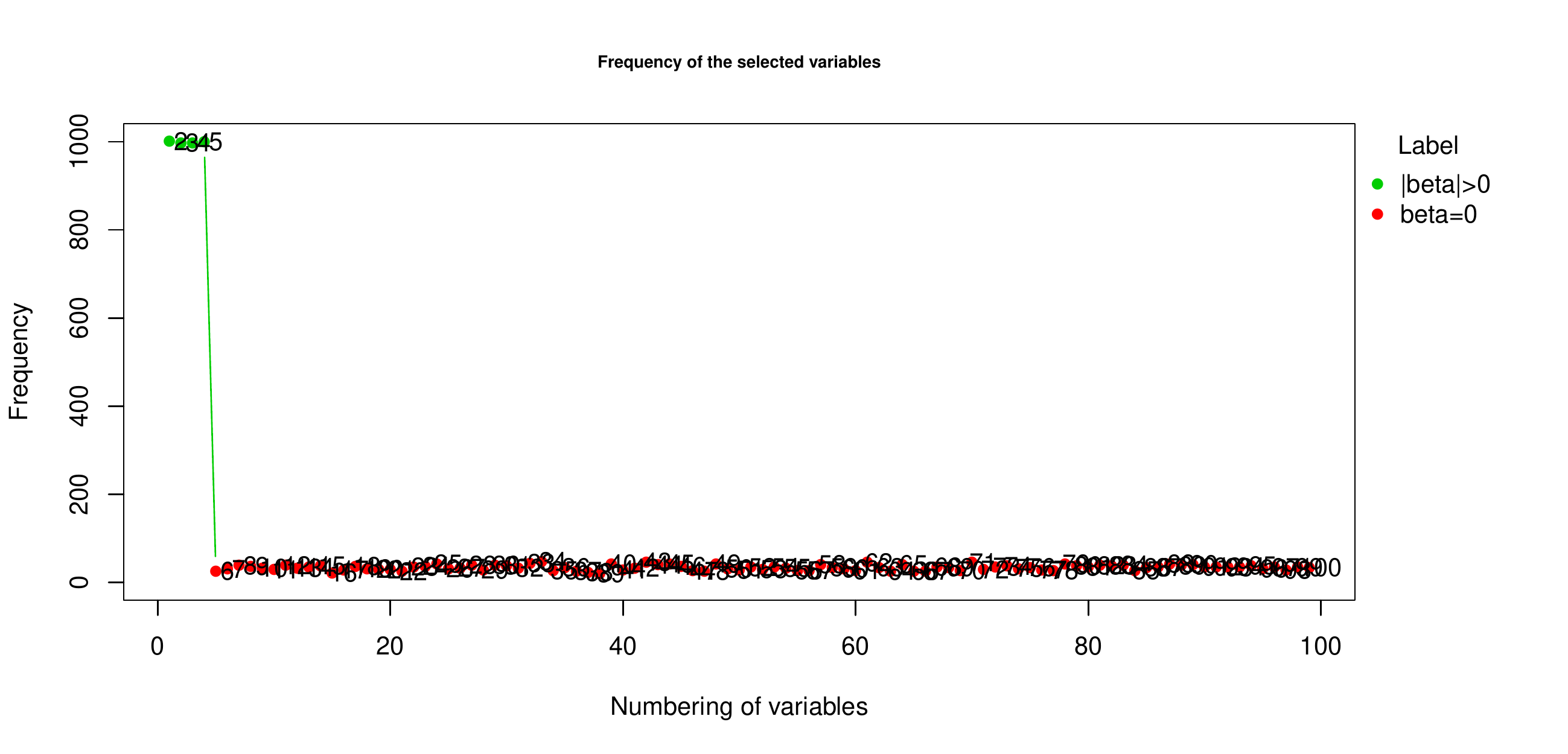}}\vspace*{-0.3cm}

   \subfigure[Case 2: ROC curve for (A)]{\includegraphics[width=0.45\textwidth,trim={0 0 0 0.5cm},clip]{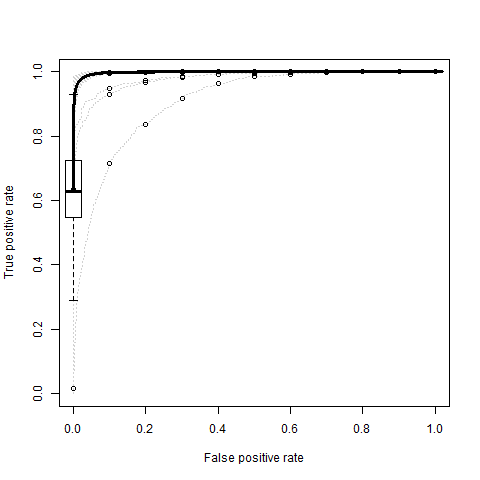}}\vspace*{-0.2cm}
   \subfigure[Case 2: ROC curve for (B)]{\includegraphics[width=0.45\textwidth,trim={0 0 0 0.5cm},clip]{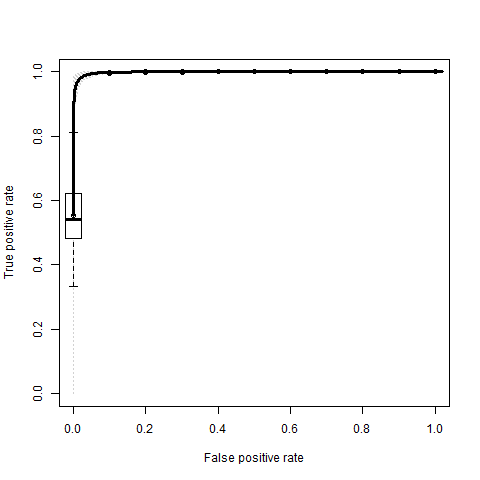}}\vspace*{-0.2cm}

   \subfigure[Case 2: ROC curve for (C)]{\includegraphics[width=0.45\textwidth,trim={0 0 0 0.5cm},clip]{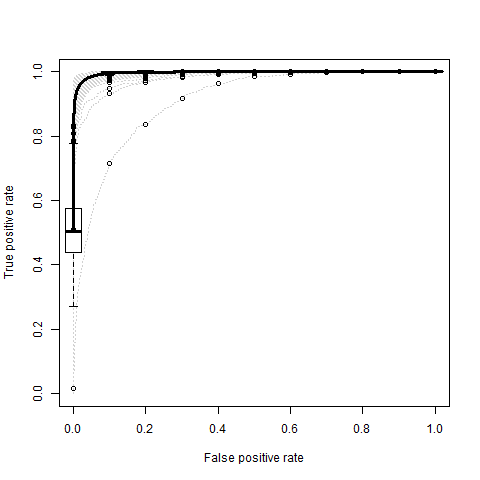}}\vspace*{-0.2cm}
   \subfigure[Case 2: ROC curve for (D)]{\includegraphics[width=0.45\textwidth,trim={0 0 0 0.5cm},clip]{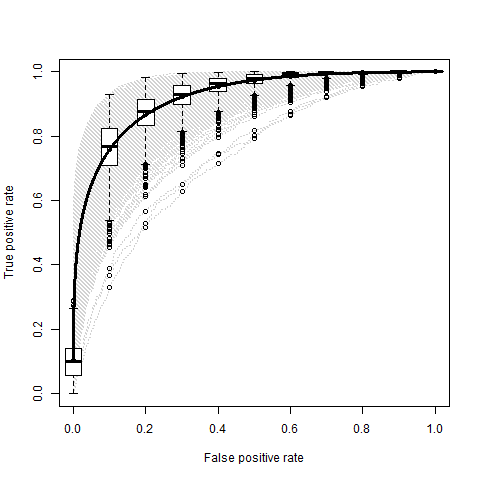}}\vspace*{-0.2cm}
\caption{The performances of the different approaches for Case 2 with 1000 replications}\label{case2_sf}

\end{figure}

\begin{figure}[t]
   \subfigure[Case 3: Variable Selection Frequencies for GATE]{\includegraphics[width=1\textwidth]{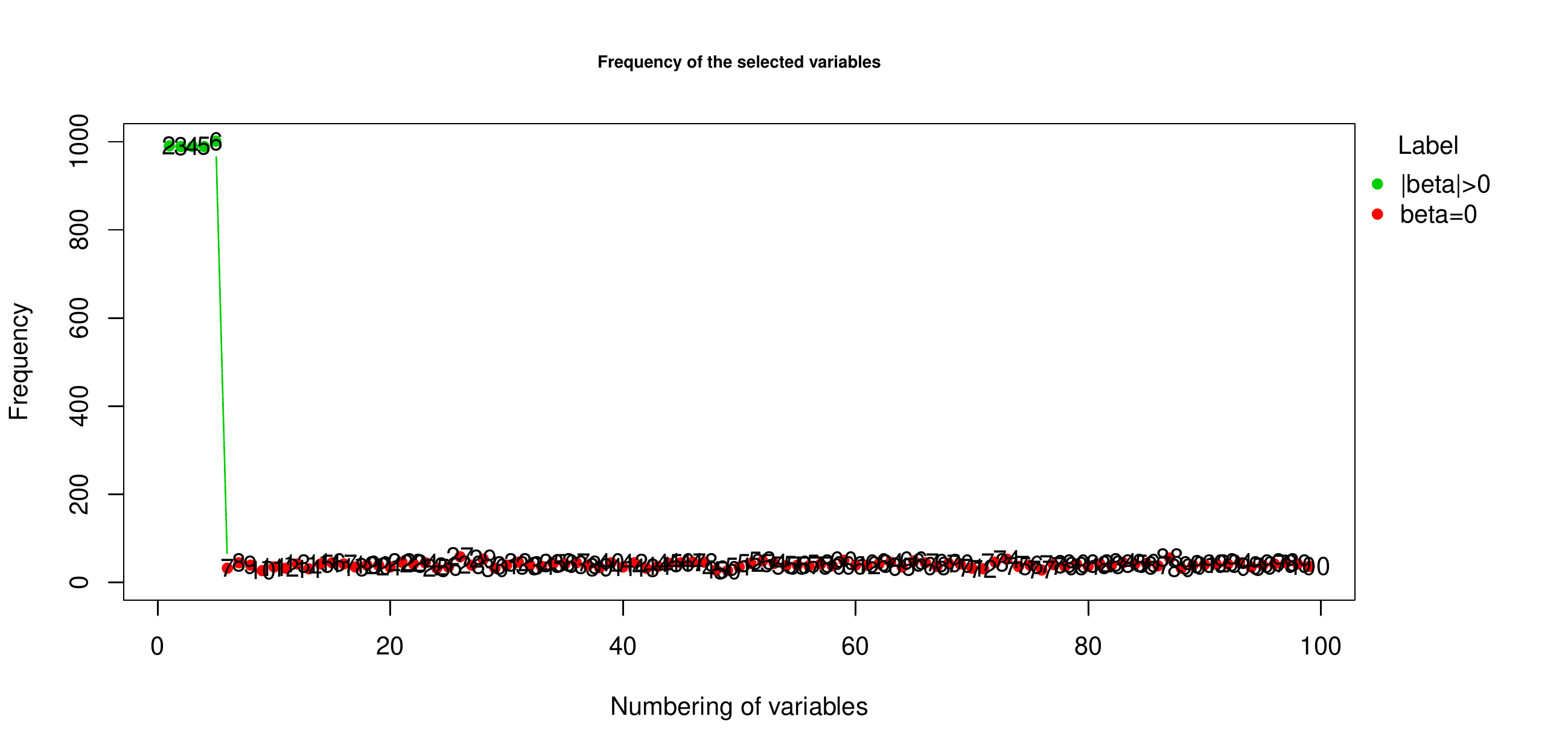}}\vspace*{-0.3cm}

   \subfigure[Case 3: ROC curve for (A)]{\includegraphics[width=0.45\textwidth,trim={0 0 0 0.5cm},clip]{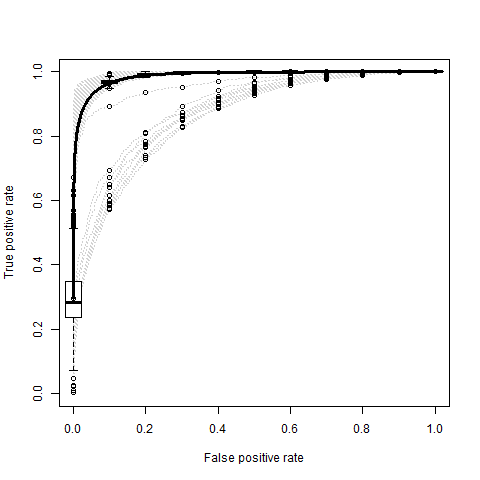}}\vspace*{-0.2cm}
   \subfigure[Case 3: ROC curve for (B)]{\includegraphics[width=0.45\textwidth,trim={0 0 0 0.5cm},clip]{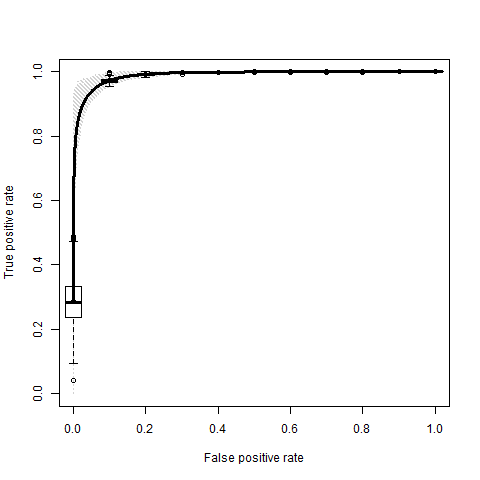}}\vspace*{-0.2cm}

   \subfigure[Case 3: ROC curve for (C)]{\includegraphics[width=0.45\textwidth,trim={0 0 0 0.5cm},clip]{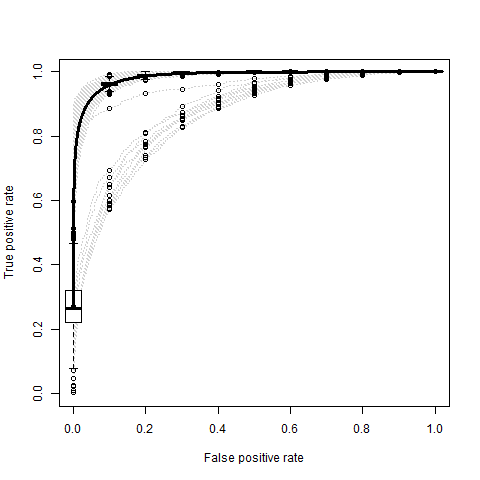}}\vspace*{-0.2cm}
   \subfigure[Case 3: ROC curve for (D)]{\includegraphics[width=0.45\textwidth,trim={0 0 0 0.5cm},clip]{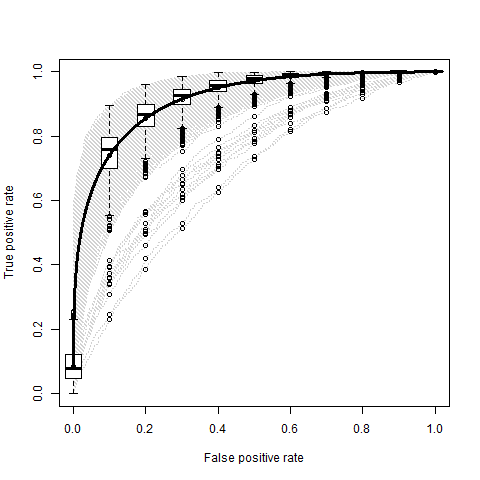}}\vspace*{-0.2cm}
\caption{The performances of the different approaches for Case 3 with 1000 replications}\label{case3_sf}
\end{figure}

In order to illustrate the advantages of the proposed active
algorithm, we compare the classification performance with {those of
the three different approaches}. The first one is to implement the
logistic classification with 15,000 training samples and 100
variables. {Second, we perform a comparison with an approach where
in each replication, following the same sample size $n$ determined
by the proposed active learning procedures, we randomly sample the
labeled point from the training set and then implement the logistic
classifier based on the same active variables identified by grafting
forward selection.} In the third approach, the 100 variables are
taken into the logistic model, and then due to the sample size in
each of our 1,000 replications, we randomly select the labeled
points from the 15,000 candidate training points for learning the
corresponding logistic classifier. Tables \ref{sim_table2},
\ref{sim_table3} and \ref{sim_table4} show the classification
results with respect to the different approaches. In addition to
summarizing the results in the tables, we also illustrate them in
Figures \ref{case1_sf}, \ref{case2_sf} and \ref{case3_sf} with
respect to the different parameter cases. In each figure, we not
only display the selection frequencies of the predictors (variables)
for our {\it GATE} learning approach but also show the 1000 ROC
curves for all replications with respect to the four different
approaches. Here, the notation (A) is used to denote our {\it GATE}
learning approach, and the notations (B), (C) and (D) are used to
denote the three comparison approaches; the scenario with whole
samples and variables; the scenario with the random subjects and
selected variables and the scenario with random subjects and whole
variables, respectively. Overall among these results,
we summarize what we find in the comparisons in the following.\\
{\bf Compared result of the scenario with whole variables and data.}
Table \ref{sim_table2} shows the classification results based on
1,000 replications for each parameter case. Here, we can treat these
results in Table \ref{sim_table2} as the baselines for comparison.
Due to these two classification measurements, ACC and AUC, we focus
on the classification results for testing sets. The lowest value of
the relative ACC of the proposed {\it GATE} approach and the
baseline values is $0.821/0.832 = 0.987$, and the lowest relative
AUC value is $0.886/0.898=0.986$. Both relative values are close to
1.000. This means that the classification performances are similar
for both approaches. However, in our {\it GATE} learning approach,
we only take less than 530 points, which is around $1/30$ of the
whole samples, and a more compact model is used in the corresponding
logistic classifier. Thus, this is the evidence to show the
efficiency of the proposed GATE learning approach, whether in the
proposed query strategy or the greedy forward selection approach.
According to the sub-figures (b) and (c) in Figures \ref{case1_sf},
\ref{case2_sf} and \ref{case3_sf}, they also show both approaches
should have similar classification results, but the results based on
the whole samples and variables, (B), should be more stable among
1,000 replications.\\
{\bf Compared result of the random subjects and fixed selected
model.} In each replication, we fix the model selected by our {\it
GATE} learning approach, and {the labeled subject set is used to
randomly sample} the subjects from the whole samples, and its size
is the same as the one we used in our active learning. Then, we
learn the corresponding logistic classifier. The results with 1,000
replications are recorded in Table \ref{sim_table3}. Consider the
classification results for the testing set. The relative values of
ACC and AUC of our {\it GATE} learning approach and this method are
all slightly larger than 1. That is, our active learning approach
still performs better in these three cases. In addition, it is
evidence in support of our greedy selection approach because we
still can achieve good classification results based on these
selected models. The sub-figures (b) and (d) in Figures
\ref{case1_sf}, \ref{case2_sf} and \ref{case3_sf} also indicate that
both approaches have almost the same classification performance.
\\
{\bf Compared result of the random subjects and whole variables.} In this classification approach, we take all 100 variables into the logistic regression model. To learn the binary classifier, we randomly select the points from the whole samples with the same sample size, $n$, for each replication. Table \ref{sim_table4} are the average classification results based on 1,000 replications. Compared with the ACC and AUC values shown in Table 1, our active learning approach significantly outperforms this random selection approach. This means that to obtain better classification results, we need to have more labeled samples if we still want to take whole variables into the model or, due to a smaller sample set, we need to have a compact logistic model. Thus, the one possible solution is our proposed active learning approach because it can be used to sequentially select the next labeled points but also identify the active variables for the binary classifier. Based on the sub-figure (e) in these three figures, the instability of this classification approach is clearly illustrated.

\begin{table}[ht]
\caption{(A) Greedy Active Learning Results}\label{sim_table1}
\scalebox{0.82}{\begin{tabular}{rlrrrrrrrrr}
\hline
    \multicolumn{3}{l}{\textcolor{blue}{ }} &\multicolumn{2}{c}{Training}&\multicolumn{2}{c}{Testing}&&Select&&
Ave. Time \\\cmidrule(r){4-5}\cmidrule(r){6-7}
Case & $p_o$    & $n$ &   ACC & AUC & ACC &AUC
 & TPR & FPR& $\#\{x_d\}$ &   (min) \\\hline
\multirow{2}{*}{1} &   \multirow{2}{*}{5} &  528.460 & 0.825 & 0.887 & 0.821 & 0.886 & 0.941 & 0.111 & 14.282 & 4.470 \\
&&(190.255)&(0.029)&(0.015)&(0.029)&(0.015)&(0.148)&(0.061)&(6.342)& \\

\multirow{2}{*}{2} &   \multirow{2}{*}{5} & 314.290 & 0.981 &  0.998 & 0.980 & 0.998 & 0.999 & 0.033 & 7.143 & 1.549 \\
&& (65.241)&(0.005)&(0.003)&(0.005)&(0.003)&(0.026)&(0.023)& (2.175) & \\
\multirow{2}{*}{3} &   \multirow{2}{*}{6} &360.250 &  0.946 & 0.987 & 0.944 & 0.987 & 0.993 & 0.040 & 8.675 & 1.950\\
&&(78.794)&(0.016)&(0.012)&(0.017)&(0.012)&(0.068)&(0.027)&(2.626)& \\ \hline
\end{tabular}}

{\scriptsize
Note: $p_o$ represents the number of non-zero parameters in the true generating model, and the standard deviation of the 1000 simulated repeats are showed in parentheses.
}
\end{table}

\begin{table}[ht]
\begin{center}
\caption{(B) Full Variables and Subjects}\label{sim_table2}
\begin{tabular}{rlrrrrr}
\hline
    \multicolumn{3}{l}{\textcolor{blue}{ }} &\multicolumn{2}{c}{Training}&\multicolumn{2}{c}{Testing} \\\cmidrule(r){4-5}\cmidrule(r){6-7}
Case & $p_o$    & $n$ &   ACC & AUC & ACC &AUC \\\hline
\multirow{2}{*}{1} &   \multirow{2}{*}{5} &  \multirow{2}{*}{15000} & 0.832 & 0.898 & 0.832 & 0.898  \\
&&&(0.025)&(0.005)&(0.025)&(0.005) \\
\multirow{2}{*}{2} &   \multirow{2}{*}{5} & \multirow{2}{*}{15000} & 0.982 &  0.999 & 0.982 & 0.999 \\
&&&(0.004)&(0.000)&(0.004)&(0.000) \\
\multirow{2}{*}{3} &   \multirow{2}{*}{6} &\multirow{2}{*}{15000} &  0.949 & 0.990 & 0.949 & 0.989 \\
&&&(0.009)&(0.001)&(0.010)&(0.002) \\ \hline
\end{tabular}
\end{center}

{\scriptsize
Note: $p_o$ represents the number of non-zero parameters in the true generating model, and the standard deviation of the 1000 simulated repeats are shown in parentheses.
}
\end{table}

\begin{table}[ht]
\begin{center}
\caption{(C) Random Subjects with the Selected Variables in (A)}\label{sim_table3}
\begin{tabular}{rlrrrrr}
\hline
    \multicolumn{3}{l}{\textcolor{blue}{ }} &\multicolumn{2}{c}{Training}&\multicolumn{2}{c}{Testing} \\\cmidrule(r){4-5}\cmidrule(r){6-7}
Case & $p_o$    & $n$ &   ACC & AUC & ACC &AUC \\\hline
\multirow{2}{*}{1} &   \multirow{2}{*}{5} &  528.460 & 0.820 & 0.885 & 0.819 & 0.884  \\
&&(190.255)&(0.029)&(0.014)&(0.029)&(0.015) \\
\multirow{2}{*}{2} &   \multirow{2}{*}{5} & 314.290 & 0.974 &  0.997 & 0.974 & 0.997 \\
&&(65.241)&(0.006)&(0.003)&(0.007)&(0.003) \\
\multirow{2}{*}{3} &   \multirow{2}{*}{6} &360.250 &  0.941 & 0.986 & 0.940 & 0.986 \\
&&(78.794)&(0.016)&(0.012)&(0.017)&(0.012) \\ \hline
\end{tabular}
\end{center}

{\scriptsize
Note: $p_o$ represents the number of non-zero parameters in the true generating model, and the standard deviation of the 1000 simulated repeats are shown in parentheses.
}
\end{table}

\begin{table}[ht]
\begin{center}
\caption{(D) Random Subjects with Full Variables}\label{sim_table4}
\begin{tabular}{rlrrrrr}
\hline
    \multicolumn{3}{l}{\textcolor{blue}{ }} &\multicolumn{2}{c}{Training}&\multicolumn{2}{c}{Testing} \\\cmidrule(r){4-5}\cmidrule(r){6-7}
Case & $p_o$    & $n$ &   ACC & AUC & ACC &AUC \\\hline
\multirow{2}{*}{1} &   \multirow{2}{*}{5} &  528.460 & 0.763 & 0.816 & 0.759 & 0.812  \\
&&(190.255)&(0.055)&(0.061)&(0.056)&(0.061) \\
\multirow{2}{*}{2} &   \multirow{2}{*}{5} & 314.290 & 0.858 &  0.922 & 0.856 & 0.921 \\
&&(65.241)&(0.030)&(0.033)&(0.031)&(0.033) \\
\multirow{2}{*}{3} &   \multirow{2}{*}{6} &360.250 &  0.851 & 0.917 & 0.848 & 0.915 \\
&&(78.794)&(0.031)&(0.034)&(0.031)&(0.034) \\ \hline
\end{tabular}
\end{center}

{\scriptsize
Note: $p_o$ represents the number of non-zero parameters in the true generating model, and the standard deviation of the 1000 simulated repeats are shown in parentheses.
}
\end{table}

\setcounter{subfigure}{0}

\setcounter{subfigure}{0}

\setcounter{subfigure}{0}

\subsection{Wave Dataset}

A well-known artificial wave dataset \citep{Breimanetal1984} is also used for illustration purposes. Originally,  there are three classes with 21 variables in this wave dataset, and the variables, of each class, are generated based on a random convex combination of two of three wave forms with noise. \citet{rak2005} expanded the dataset by adding noise variables and generating more subjects. He also modified this dataset as a binary response dataset by only keeping the subjects of the first two classes. This binary classification dataset can be downloaded from the following link,
\url{http://eric.univ-lyon2.fr/\~ricco/tanagra/fichiers/wave_2_classes_with_irrelevant_attributes.zip}.

Currently there are 33,334 subjects in this dataset, and 10,000 subjected are collected as the training set. It is a balanced classification problem, because the subjects numbers in both classes are almost the same.
In addition to the original 21 predictors, 100 noise variables are added, and they are completely independent from the corresponding classification problem.
\citet{rak2005} analyzed this dataset via a free data mining software, ``TANAGRA'', and he not only used logistic regression for the classification problem but also performed forward selection according to a SCORE test based on the whole training set. Overall, the classification error rate for testing set was $7.87\%$. For the variable selection results, no noise variables are selected into the classification model, and 15 variables from the original 21 active variables are identified as active variables by setting the significant level as $1\%$ in the SCORE test. For more details, please refer to the following website: \url{http://data-mining-tutorials.blogspot.tw/2008/12/logistic-regression-software-comparison.html}.

To implement the proposed {\it GATE} learning approach, we choose the same tuning parameters as those used for simulation studies in Section 3.1. That is, the size of the initial set is 100 points, the order for the subject candidate is $h = 200$, and the batch size is also chosen as 30. For the stopping criterion, we still set the threshold value, $\varepsilon$, as $10^{-2}$.
Here, we repeat our active learning procedure 100 times by randomly generating the initial design set from the training set and treat the remaining points in the training set as the unlabeled points. In addition to the proposed approach, we also implement the other three types of methods as shown in Section 3.1. The overall performances of these four approaches are summarized in Table \ref{art_example}. Basically, the proposed {\it GATE} approach selects an average of 743 samples to learn the classification model. Then, based on the testing set, the average value of the classification rate is 0.920, and the mean value of the AUC value among 100 replications is 0.980. Compared with the classification error rate of TANAGRA, our classification result is quite good, in particular when we only use less than $7.5\%$ of the entire training samples. For the other three approaches, first, the logistic classification procedure with the whole predictors and the training set has identical results to those of TANAGRA, and it can only be implemented one time. With respect to the other two methods, logistic classification for the cases of the random subjects and selected variables and the random subject with full variables, their performances are similar to those in the simulation studies. Overall, the case with random subjects and whole variables has the worst performances in terms of ACC and AUC. This should also be related to the fact that the subject size is too small compared to the model size, and thus the uncertainty of the classification model increases. When we fixed the variables as the selected variables, the performance of this case was similar to those of our proposed {\it GATE} learning procedure.

\begin{figure}[t]
\begin{center}
\includegraphics[width=12cm]{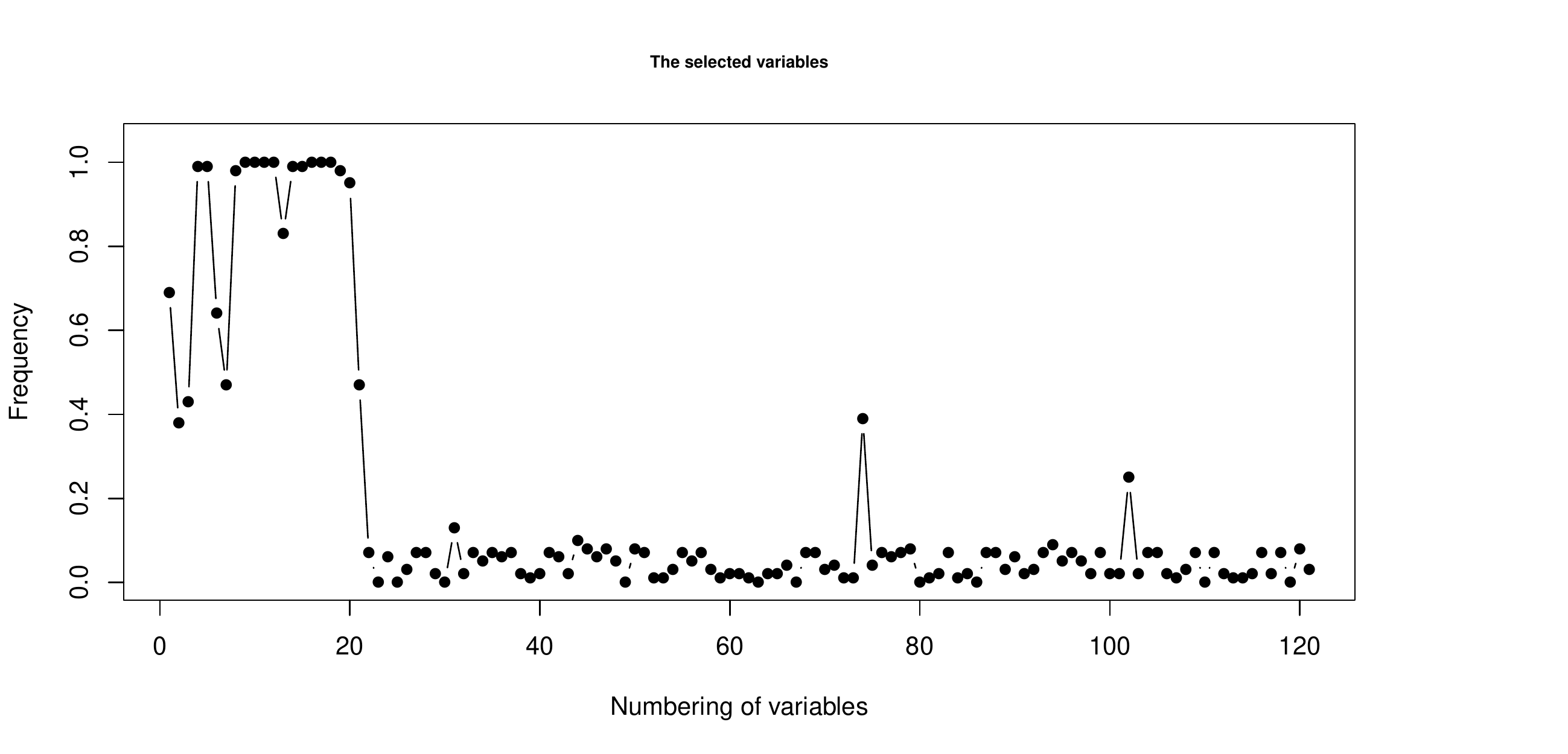}
\end{center}
\caption{The variable selection frequencies for the artificial example.} \label{sel_art}
\end{figure}

It is of interest to investigate the variable selection results.
Given the whole training set, TANAGRA can identify the important variables via the forward selection procedure according to the SCORE test. There are 15 variables, $V_4, V_5, V_8, V_9, V_{10}, V_{11},  V_{12},$  $V_{13},  V_{14},$ $V_{15},$  $V_{16},$ $V_{17},$ $V_{18},$ $V_{19}$ and $V_{20}$, identified as the important variables.
Consider the proposed {\it GATE} learning procedure.
The selected frequencies of each variable are shown in Figure \ref{sel_art}. Suppose we set 0.8 as a threshold value for the selected frequencies. Then, it is clear that our grafting approach can also yield the same set of active variables. For these selected 15 variables, except $V_{13}$, the selected frequencies are higher than and equal to 95\%. In fact, the frequency at which $V_{13}$ is identified is at least 83\%. For 100 noise variables, the selected frequencies are essentially lower than those of the first 21 original variables.

\begin{table}[t]
\begin{center}
\caption{The Results of the Artificial Example}\label{art_example}
\scalebox{0.82}{\begin{tabular}{lrrrrrr} \hline
    \multicolumn{2}{l}{\textcolor{blue}{ }} &\multicolumn{2}{c}{Training}&\multicolumn{2}{c}{Testing}& Variable Size \\\cmidrule(r){3-4}\cmidrule(r){5-6}
Method  & $n$ &   ACC & AUC & ACC & AUC &$\#\{x_d\}$ \\\hline
\multirow{2}{*}{(A) GATE Learning} &  743.500 & 0.920 & 0.978 & 0.920 & 0.980 & 22.45  \\
&(444.264)&(0.002)&(0.001)&(0.002)&(0.001)&(14.809)\\
\hline
(B) Full Variables  & 10000 & & & 0.921& 0.980 &121\\
and Subjects& & & & & & \\ \hline
(C) Random Subjects   & 743.500 & & & 0.917& 0.978 &22.45\\
with Selected Variables&(444.264) & & &(0.003) &(0.001) &(14.809) \\
\hline
(D) Random Subjects  & 743.500 & & & 0.857& 0.937 &121\\
with Full Variables &(444.264) & & &(0.019) &(0.014) &  \\ \hline
\end{tabular}}
\end{center}
\end{table}

\section{Discussion and Conclusion}

In this study, we propose a logistic model-based active learning
procedure for binary response data named {\it GATE} algorithm. In
addition to the common subject selection feature in active learning
procedures, our algorithm can also identify the proper
classification model with the given data. We propose a two-stage
subject selection procedure combining the ideas of uncertainty
sampling and locally $D$-optimal criterion from the experimental
design methods. In our algorithm, we use {the} grafting approach,
{a} greedy forward selection procedure, and adopt a sequential batch
selection strategy at each stage. Hence, the proposed active
learning algorithm will repeat the subject selection and forward
variable selection steps until a stopping criterion is fulfilled.
Both numerical results with our synthesized data and a well-known
artificial example support the success of the proposed method.
{To use our {\it GATE} algorithm, we need to specify (1) the batch
size $n_q$, (2) maximum tolerance distance,  (3) uncertainty sample
step $h$, and (4) threshold value in the stopping criterion
$\varepsilon$.} To select the most suitable parameters with a given
data set is always an important and time-consuming issue. Because we
consider subject and variable selection simultaneously, some
redundant variables might be added to the current classification
model due to the current selected subjects. The proposed algorithm
tends to over-select variables since this situation is common in the
forward selection-based procedures. A natural approach to avoid this
situation is to consider a step-wise selection approach instead of a
simple forward selection procedure. Once users adopt this step-wise
approach, then they will have chances to remove redundant variables
from the current model at backward elimination step. Certainly, the
computational time will be largely extended for this kind of
approaches, thus users should take this extra computational cost
into consideration for choosing the most proper approach for their
applications.

\section*{Acknowledgments}

The research of Chen was partially supported by the National Science
Council under Grant MOST105-2628-M-006-002-MY2 and the Mathematics
Division of the National Center for Theoretical Sciences in Taiwan.


\section*{Reference}

\begin{thebibliography}{}

\bibitem[Atkinson, 1996]{Atkinson96}
Atkinson, A.~C. (1996).
\newblock The usefulness of optimum experimental designs.
\newblock {\em Journal of the Royal Statistical Society. Series B
  (Methodological)}, 58(1):59--76.

\bibitem[Breiman et~al., 1984]{Breimanetal1984}
Breiman, L., Friedman, J., Olshen, R., and Stone, C. (1984).
\newblock Classification and regression trees.
\newblock {\em Belmont, CA: Wadsworth Statistical Press}.

\bibitem[Cohn et~al., 1994a]{Cohn94a}
Cohn, D., Atlas, L., and Ladner, R. (1994a).
\newblock Improving generalization with active learning.
\newblock {\em Machine Learning}, 15(2):201--221.

\bibitem[Cohn, 1996]{Cohn96}
Cohn, D.~A. (1996).
\newblock Neural network exploration using optimal experiment design.
\newblock {\em Neural Netw.}, 9(6):1071--1083.

\bibitem[Cohn et~al., 1994b]{Cohn94}
Cohn, D.~A., Ghahramani, Z., and Jordan, M.~I. (1994b).
\newblock Active learning with statistical models.
\newblock In {\em Proceedings of the 7th International Conference on Neural
  Information Processing Systems}, NIPS'94, pages 705--712, Cambridge, MA, USA.
  MIT Press.

\bibitem[Culver et~al., 2006]{Culver06}
Culver, M., Kun, D., and Scott, S. (2006).
\newblock Active learning to maximize area under the roc curve.
\newblock In {\em Sixth International Conference on Data Mining, 2006. ICDM
  '06.}

\bibitem[Deng et~al., 2009]{Dengetal2009}
Deng, X., Joseph, V.~R., Sudjianto, A., and Wu, C. F.~J. (2009).
\newblock Active learning through sequential design, with applications to
  detection of money laundering.
\newblock {\em Journal of the American Statistical Association},
  104(487):969--981.

\bibitem[Efron and Hastie, 2016]{Efron2016}
Efron, B. and Hastie, T. (2016).
\newblock {\em Computer Age Statistical Inference: Algorithms, Evidence, and
  Data Science}.
\newblock Cambridge University Press, New York, NY, USA, 1st edition.

\bibitem[Fedorov, 1972]{Fedorov1972}
Fedorov, V. (1972).
\newblock {\em Theory Of Optimal Experiments}.
\newblock Probability and Mathematical Statistics. Elsevier Science.

\bibitem[Friedman et~al., 2010]{Friedman10}
Friedman, J., Hastie, T., and Tibshirani, R. (2010).
\newblock Regularization paths for generalized linear models via coordinate
  descent.
\newblock {\em Journal of Statistical Software}, 33(1):1--22.

\bibitem[Hsu, 2010]{hsu10}
Hsu, D.~J. (2010).
\newblock {\em Algorithms for Active Learning}.
\newblock PhD thesis, Columbia University.

\bibitem[Kubicaa et~al., 2011]{kubicaaetal2011}
Kubicaa, J., Singhb, S., and Sorokinac, D. (2011).
\newblock Parallel large-scale feature selection.
\newblock {\em Scaling up Machine Learning: Parallel and Distributed
  Approaches}.

\bibitem[Lewis and Gale, 1994]{Lewis94}
Lewis, D.~D. and Gale, W.~A. (1994).
\newblock A sequential algorithm for training text classifiers.
\newblock In {\em Proceedings of the 17th Annual International ACM SIGIR
  Conference on Research and Development in Information Retrieval}, SIGIR '94,
  pages 3--12, New York, NY, USA. Springer-Verlag New York, Inc.

\bibitem[Long et~al., 2010]{Long2010}
Long, B., Bian, J., Chapelle, O., Zhang, Y., Inagaki, Y., and Chang,
Y. (2010).
\newblock Active learning for ranking through expected loss optimization.
\newblock In {\em SIGIR 2010}.

\bibitem[Mallat and Zhang, 1993]{Mallat_93}
Mallat, S.~G. and Zhang, Z. (1993).
\newblock Matching pursuits with time-frequency dictionaries.
\newblock {\em IEEE Transactions on Signal Processing}, 41(12):3397--3415.

\bibitem[Perkins et~al., 2003]{Perkinsetal2003}
Perkins, S., Lacker, K., and Theiler, J. (2003).
\newblock Grafting: Fast, incremental feature selection by gradient descent in
  function space.
\newblock {\em The Journal of Machine Learning Research}, 3:1333--1356.

\bibitem[Rakotomalala, 2005]{rak2005}
Rakotomalala, R. (2005).
\newblock Tanagra: a free software for research and academic purposes.
\newblock In {\em Proceedings of European Grid Conference 2005, RNTI-E-3},
  volume~2, pages 697--702.

\bibitem[Settles, 2009]{settles.tr09}
Settles, B. (2009).
\newblock Active learning literature survey.
\newblock Computer Sciences Technical Report 1648, University of
  Wisconsin--Madison.

\bibitem[Settles, 2011]{settles11}
Settles, B. (2011).
\newblock From theories to queries: Active learning in practice.
\newblock {\em Journal of Machine Learning Research, Workshop on Active
  Learning and Experimental Design,}, Workshop and Conference Proceedings 16:1
  -- 18.

\bibitem[Settles, 2012]{Settle2012}
Settles, B. (2012).
\newblock {\em Active learning. Synthesis lectures on artificial intelligence
  and machine learning}.
\newblock Morgan \& Claypool Publishers, San Rafael.

\bibitem[Silvey, 1980]{Silvey80}
Silvey, S.~D. (1980).
\newblock {\em Optimal design : an introduction to the theory for parameter
  estimation}.
\newblock London ; New York : Chapman and Hall.

\bibitem[Singh et~al., 2009]{singhetal2009}
Singh, S., Kubica, J., Larsen, S., and Sorokina, D. (2009).
\newblock Parallel large scale feature selection for logistic regression.
\newblock In {\em SDM}, pages 1172--1183. SIAM.

\bibitem[Whitney, 1971]{whiteny1971}
Whitney, A.~W. (1971).
\newblock A direct method of nonparametric measurement selection.
\newblock {\em IEEE Trans. Comput.,}, 20(9):1100--1103.

\bibitem[Yuan et~al., 2012]{Yuan:2012}
Yuan, G.-X., Ho, C.-H., and Lin, C.-J. (2012).
\newblock An improved glmnet for l1-regularized logistic regression.
\newblock {\em J. Mach. Learn. Res.}, 13(1):1999--2030.

\end{thebibliography}

\end{document}